\documentclass[11pt,onecolumn]{article}         
\usepackage[top=3.6cm, bottom=3.2cm, left=2.3cm, right=2.3cm]{geometry}

\usepackage{color}
\usepackage{graphicx}
\usepackage{cite}
\usepackage[cmex10]{amsmath}
\usepackage{amssymb}
\usepackage{subfig}
\usepackage{multirow}
\usepackage{makecell}
\usepackage[colorlinks=true,linkcolor=red,citecolor=blue]{hyperref}
\usepackage{footmisc}
\usepackage{algorithmic}
\usepackage[ruled,vlined]{algorithm2e}
\graphicspath{{images/}}
\begin{document}

\title{Learning the Synthesizability of Dynamic Texture Samples\footnote{The project website is available at \url{http://captain.whu.edu.cn/project/DTsynthesizability.html}.}}

{
\author{Feng~Yang$^{1}$, Gui-Song~Xia$^{1}$, Dengxin~Dai$^{2}$, Liangpei~Zhang$^1$\\
\\
$^1${\em State Key Lab. LIESMARS, Wuhan University, China}\\
{\tt \small \{guisong.xia, fengyang, zlp62\}@whu.edu.cn}\\
$^2${\em Computer Vision Lab., ETH Zurich, Switzerland}\\
{\tt \small dai@vision.ee.ethz.ch}\\
}
}

\maketitle

\begin{abstract}
A dynamic texture (DT) refers to a sequence of images that exhibit temporal regularities and has many applications in computer vision and graphics. Given an exemplar of dynamic texture, it is a dynamic but challenging task to generate new samples with high quality that are perceptually similar to the input exemplar, which is known to be {\em example-based dynamic texture synthesis (EDTS)}.
Numerous approaches have been devoted to this problem, in the past decades, but none them are able to tackle all kinds of dynamic textures equally well. In this paper, we investigate the synthesizability of dynamic texture samples : {\em given a dynamic texture sample, how synthesizable it is by using EDTS, and which EDTS method is the most suitable to synthesize it?}
To this end, we propose to learn regression models to connect dynamic texture samples with synthesizability scores, with the help of a compiled dynamic texture dataset annotated in terms of synthesizability.
More precisely, we first define the synthesizability of DT samples and characterize them by a set of spatiotemporal features. Based on these features and an annotated dynamic texture dataset,
we then train regression models to predict the synthesizability scores of texture samples, and learn classifiers to select the most suitable EDTS methods.
We further complete the selection, partition and synthesizability prediction of dynamic texture samples in a hierarchical scheme. We finally apply the learnt synthesizability to detecting synthesizable regions in videos. The experiments demonstrate that our method can effectively learn and predict the synthesizability of DT samples.
\end{abstract}

\section{Introduction}
Dynamic patterns, such as running water, traffic flow, fire and smoke and so on, are a set of commonly seen visual phenomena in natural world and have attracted many attentions in computer vision, named after dynamic textures or temporal textures~\cite{doretto2003dynamic,yuan2004synthesizing,costantini2008higher,schodl2000video,kwatra2003graphcut,yang2016stationary}. By the term of ``dynamic textures'' (DTs) in this paper, we refer to any spatial-temporal process as sequence of images that exhibit temporal regularities, \emph{i.e.}, certain stationarity properties in time~\cite{doretto2003dynamic}. A substantial amount of work has been devoted to synthesizing dynamic textures from examples, \emph{i.e.}, example-based dynamic texture synthesis (EDTS)~\cite{wei2009state}, which aims to generate new dynamic texture of desired length or size that is perceptually similar to the DT example. Recent years have witnessed significant progress in example-based dynamic texture synthesis algorithms, either in parametric models~\cite{doretto2003dynamic,doretto2004spatially,yuan2004synthesizing,costantini2008higher,XiaFPA12,xia2014synthesizing,
xie2016synthesizing} or non-parametric models~\cite{schodl2000video,wei2000fast,kwatra2003graphcut,bar2001texture,yang2016stationary}. While many attempts mainly focused on synthesis approaches to generate high-quality results, how to analyse and evaluate DTs as input samples for synthesis has not been addressed yet.

\subsection{Problem statement}
The goal of example-based dynamic texture synthesis (EDTS) can be stated as follows: given a dynamic texture sample, synthesize a novel video sequence that looks perceptually equivalent to the input sample with similar appearance and dynamics.
This task reveals that the result of generated dynamic texture is subject to both synthesis algorithm and input exemplar, where the former promises technical support and the latter provides material to be simulated.
We take interest in dynamic texture synthesizability with respect to both DT samples and EDTS methods - how to analyse and evaluate DTs as input samples to aid EDTS. Our studies of learning synthesizability of dynamic texture samples are based on following observations:
\begin{itemize}
\item[-] \emph{No existing EDTS methods can tackle all kinds of DTs equally well.}
Although a growing number of approaches have emerged in EDTS to generate dynamic textures of high quality, none of the methods can synthesize all dynamic textures equally well. Each method has its own benefits and limits. For instance, statistic parametric models~\cite{doretto2003dynamic,costantini2008higher,xia2014synthesizing} are mathematically sound, but likely to suffer from degraded visual quality because the statistics are not complete enough or not well enforced. And the non-parametric models such as patch-based methods~\cite{wei2000fast,kwatra2003graphcut} though efficient sometimes produce verbatim patterns. Moreover, some methods~\cite{doretto2004spatially,xia2014synthesizing} are good at stationary dynamic texture synthesis in joint spatial-temporal domain, while others~\cite{doretto2003dynamic,costantini2008higher} do better in control synthesis along time axis.

\item[-] \emph{Not all dynamic textures can be equally well reproducible in synthesis.}
On the other hand, dynamic texture samples are not equally well reproducible. Except when dealing with a controlled video acquisition protocol, for instance in specific lab environments, DT samples usually contain outliers, cluttered backgrounds, non-uniform illumination, or even objects and complex scenes. If not at a fixed viewpoint, video sequences of DTs are captured unstably by cameras with panning or jittering.
The appearance of DTs can vary largely, ranging from fire and smoke to crowds and traffic flow.
Dynamics also exhibit different motion patterns, e.g. dominant orientation, heterogeneous orientation and stochastic dynamics.
The diversities of appearance and dynamics in various DTs bring difficulties and challenges in EDTS.
It would be helpful to evaluate dynamic texture samples input for EDTS according to how well its appearance and dynamics can be reproduced by only analysing the original sample.

\item[-] \emph{How to suggest suitable EDTS methods for DT samples?}
Neither are existing synthesis methods able to tackle all kinds of dynamic textures, nor can dynamic textures be equally well reproducible.
It is difficult to find a universal approach to synthesize all kinds of DTs well. Since many existing EDTS methods are available, why not choose a flexible strategy to take advantage of their strengths in synthesis.
Can we suggest appropriate EDTS methods adapted to different DT samples? Rather than struggle to build a perfect EDTS method which is extremely difficult to implement, we select one case-optimal method among several existing alternatives, in order to reach success rates better than those of any individual method. 
It is a comprehensive way to draw on the wisdom of the masses.
For instance, linear dynamic system (LDS)~\cite{doretto2003dynamic} is a statistical generative model for synthesizing dynamic textures, but will have a tendency toward smoothing the dynamics and degrading visual quality over time. It would be beneficial that one could resort to an alternative method if in such case when long sequences need to be synthesized.
This leads to a win-win situation both to DT samples and EDTS methods: provide good DT samples for EDTS methods, and choose suitable EDTS methods for DT samples.
\end{itemize}

\subsection{Motivation and objective}

Although it is intuitive that some videos will be easier to synthesize than others, quantifying this intuition has not been addressed in previous studies.
There are no previous investigations that try to quantify individual DTs in terms of how synthesizable they are, and few computer vision systems to predict synthesizability and suggest suitable synthesized methods for DT samples. Also, there are no databases of videos calibrated in terms of the degree of synthesizability for each dynamic texture.
Induced by these issues, we are going to investigate the synthesizability of dynamic texture - how well its underlying dynamic patterns can be reproduced by only analyzing the original sample.

\begin{figure}[htb!]
    \centering
    \subfloat[\normalsize Prediction of synthesizability.]{
     \label{fig:pre_synthesizability}
    \includegraphics[width = 0.48\linewidth]{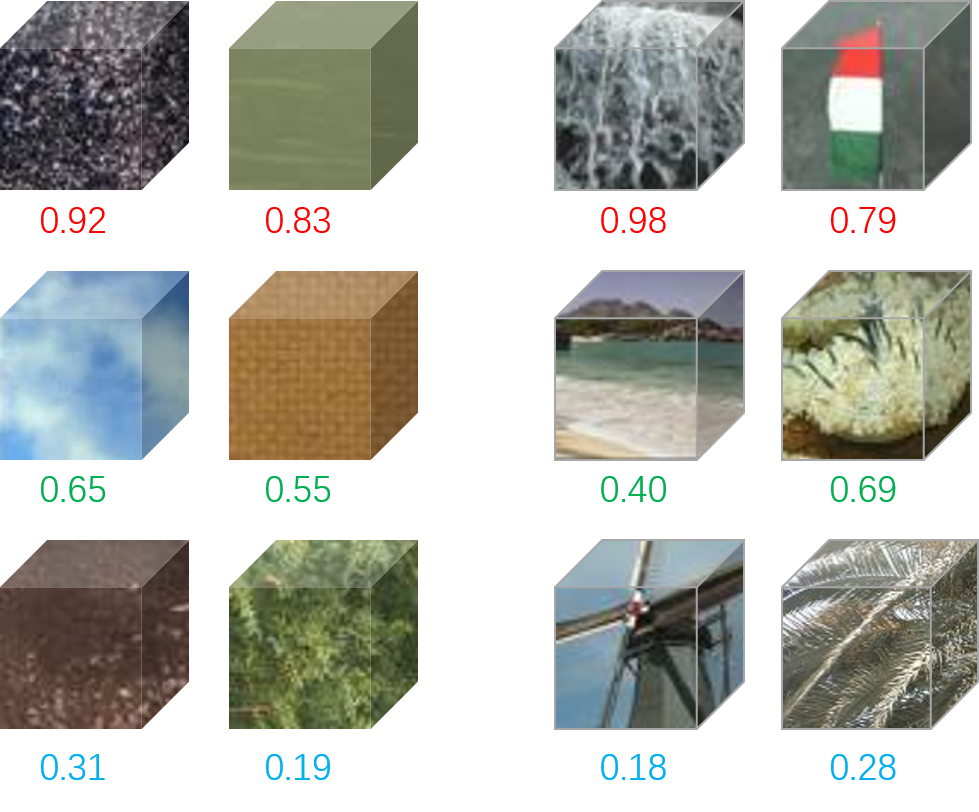}
 }

    \centering
    \subfloat[\normalsize Synthesizable regions detection for synthesis.]{
     \label{fig:detect_region}
    \includegraphics[width = 0.5\linewidth]{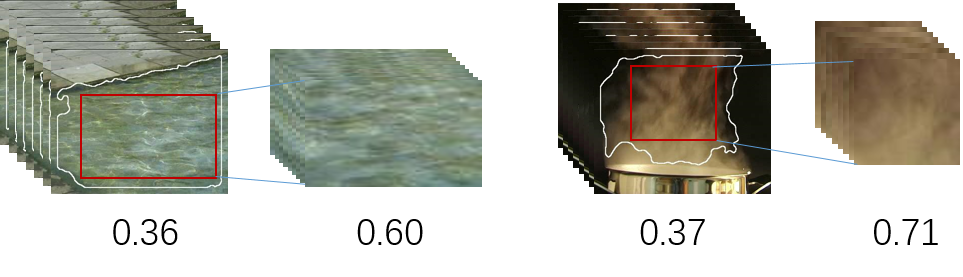}
 }
   \caption{(a) Synthesizability of dynamic texture samples predicted by our approach. The values are in $[0, 1]$ and a higher value indicates the example is easier to synthesize. Left: SHDTs, right: TDTs. (b) We also detect the most synthesizable region to trim videos into rectangular homogeneous regions.}
\label{fig:examples}
\vspace{-3mm}
\end{figure}

We characterize the synthesizability of a dynamic texture sample as the probability that existing EDTS methods will produce good synthesized results for a specific DT sample, prior to using any synthesis method to synthesize it.
The seminal work to predict synthesizability of static texture is proposed by Dai {\em et al.}~\cite{dai2014synthesizability}.
Inspired by their work, we propose in this paper to predict synthesizability score of a given dynamic texture sample, and suggest which EDTS method is best suited to synthesize it.
Fig.~\ref{fig:pre_synthesizability} shows the synthesizability scores assigned to some dynamic texture samples by our system.
Although dynamic textures are common in natural scenes, in many cases only a part of the scene forms a dynamic texture. It would be useful to tailor videos to regions with good synthesizability by discarding undesirable background. Fig.~\ref{fig:detect_region} illustrates the trimming of video into its most synthesizable and rectangular homogeneous regions (the red boxes).

In order to learn the synthesizability of dynamic textures, we have collected a dataset of dynamic textures with manually annotated synthesizability scores according to the synthesis results by a set of EDTS methods. We connect dynamic texture samples with synthesizability scores via a learning scheme, where regression models are learned from the collection of annotated data.
Feature representation is imperative for learning synthesizability. We designed a SCOP-DT descriptor specific for dynamic texture representation, which extended the \emph{shape-based co-occurrence patterns} (SCOP)~\cite{Xia2017Texture} from 2D static texture to 3D dynamic texture by incorporating temporal cues implicitly.

\subsection{Contributions}
Our work is distinguished in following aspects:
\begin{itemize}
\item[-] We make it possible to estimate the synthesizability of dynamic textures by learning regression models with appropriate spatiotemporal descriptors. Meanwhile, for a given DT sample, our system can also suggest the ``best'' synthesis method from off-the-shelf EDTS algorithms.
\item[-] To the best of our knowledge, we first investigate ``dynamic textureness'' property of dynamic patterns in video to automatically discern dynamic textures, and divide dynamic textures into two subcategories based on the spatial modes to suit different synthesis capacity.
\item[-] We proposed a novel SCOP-DT descriptor for dynamic texture representation, which can capture geometrical aspects and temporal consistency simultaneously.
\item[-] We compile a dynamic texture dataset with synthesizability annotations.
\end{itemize}

The rest of the paper is organized as follows. Section~\ref{RW} briefly recalls related work. Section~\ref{sec:synthesizability} presents the problem formulation of learning dynamic texture synthesizability. Section~\ref{sec:method} investigates dynamic texture representation relevant to synthesizability and depicts the proposed method of learning synthesizability. Section~\ref{sec:dataset} introduces our dataset collected for learning synthesizability. Section~\ref{sec:experiment} demonstrates the experimental results and analysis. Finally, Section~\ref{sec:conclude} draws some conclude remarks.
Note that all the experimental results are available at \url{http://captain.whu.edu.cn/project/DTsynthesizability.html}.

\section{Related work}
\label{RW}
This section briefly recalls some researches that are closely related to our work.

\paragraph{Example-based dynamic texture synthesis:}
The synthesizability aims to help find good samples and suggest suitable EDTS methods. Recent years have witnessed significant progress in example-based texture synthesis algorithms, roughly divided into two main categories of approaches: parametric and non-parametric methods.
Parametric methods usually define a parametric model consisting of a set of statistical measurements that cover the spatial extent and temporal domain of dynamic textures, e.g. spatiotemporal autoregressive (STAR) model~\cite{szummer1996temporal}, linear dynamic system (LDS) model~\cite{doretto2003dynamic} and its variants~\cite{yuan2004synthesizing,woolfe2006shift,chan2007classifying,costantini2008higher}. While most of LDS-based methods are subject to learning the temporal statistics, Doretto {\em et al.}~\cite{doretto2004spatially}  tried to jointly model the spatio-temporal statistics by dynamic multiscale autoregressive models. Xia {\em et al.}~\cite{xia2014synthesizing,XiaFPA12} developed a compact Gaussian texton representation to synthesize stationary dynamic textures. Based on convolutional neural network, a spatial-temporal generative ConvNet (STGConvNet)~\cite{xie2016synthesizing} has been proposed to model and synthesize dynamic textures. Parametric methods build explicit models with mathematic foundation, but the main challenge lies in designing rigid and meaningful mathematical models that are able to capture the essence of different dynamic patterns. Due to the incomplete or unenforced statistics, parametric models often failed to synthesize complex geometry patterns.

Rather than build explicit models with parameters estimated, non-parametric methods often bypass modeling spatial-temporal mathematic mechanism of dynamic texture, mainly including copy-based methods and feature-oriented synthesis. Copy-based methods produce new dynamic textures by resampling small parts from an input sample as elements, to synthesize in spatiotemporal domain~\cite{wei2000fast,kwatra2003graphcut} or along time~\cite{schodl2000video}.
Though efficient and visual results are strikingly good, copy-based methods likely lead to verbatim reproduction, too close to a mere copy-paste. Methods of feature-oriented synthesis match statistics of features between synthetic videos and original ones~\cite{bar2001texture,yang2016stationary}. This subgroup of methods do not generate verbatim repetition, but need to carefully choose or design statistics and spatiotemporal features.

\paragraph{Image or video quality evaluation:}
Several works have investigated quantifying certain qualitative characteristics of image or video in computer vision. These include interestingness~\cite{gygli2013interestingness}, memorability~\cite{isola2011makes}, quality~\cite{luo2008photo} or city geo-awareness~\cite{doersch2012what}, which leverage data-driven approaches to compute high-level visual attributes likely concerned with psychological perception.
The most related work to ours is proposed by Dai {\em et al.}~\cite{dai2014synthesizability} which predicted synthesizability of static texture. However, the question of learning the synthesizability of dynamic texture remains an open issue. We introduce synthesizability into dynamic textures and develop a series of approaches in a dynamical setting.

\paragraph{Dynamic texture recognition:}
Our work is also related to dynamic texture recognition in the sense of representation and classification. For recognition, it includes several major procedures, like feature extraction~\cite{zhao2007dynamic_pami,derpanis2012spacetime}, metric learning~\cite{ghanem2010maximum,ravichandran2013categorizing}, classifier design~\cite{ghanem2010sparse}. In our task, we also need to exploit spatiotemporal features for dynamic texture representation. 
\section{Dynamic Texture Synthesizability}\label{sec:synthesizability}
\subsection{Problem formulation}
Denote $V \in \mathbb R^{U \times d}$ as a video with $d$ channels defined in space-time domain $U = \{0,\dots,H-1\} \times \{0,\dots,W-1\} \times \{0,\dots,T-1\}$. In particular, $d = 1$ for grey-scale videos and $d = 3$ for color videos.
The synthesizability $s \in [0, 1]$ of a DT sample $V$ indicates the probability that EDTS methods will produce good synthesized results for $V$ before using any synthesis method to synthesize it. A DT sample $V$ with synthesizability score $s$ is denoted as $(V, s)$.

\begin{figure}[ht!]
  \centering
  \includegraphics[width= 0.9\linewidth]{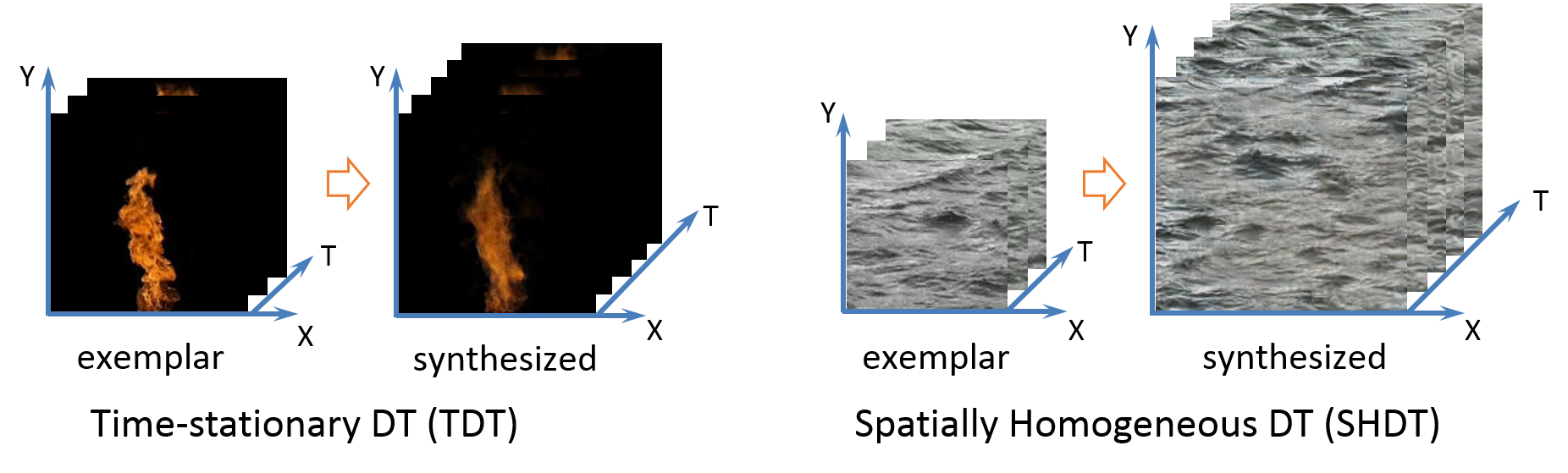}
\vspace{-3mm}
   \caption{An example to show the difference between time-stationary dynamic texture (TDT) and spatially homogeneous dynamic texture (SHDT).}
   \label{fig:TDT_SHDT}
\vspace{-1mm}
\end{figure}

\emph{SHDT and TDT:} Dynamic textures are spatial-temporal visual patterns that exhibit temporal regularities, but do not necessarily show statistical stationarity in the spatial domain. If imposed by spatially stationary constraints as well, we call these \emph{spatially homogeneous dynamic textures} (SHDTs)~\cite{doretto2004spatially}. To differentiate, for the DTs only showing temporal stationarity, we refer to \emph{time-stationary dynamic textures} (TDTs). Synthesis of TDTs can be only conducted along time axis, whereas it is potential to synthesize SHDTs in both the spatial and temporal domains. See differences in Fig.~\ref{fig:TDT_SHDT} for example.
Correspondingly, spatial and temporal synthesizability can be pre-computed for every SHDT sample, while TDTs merely cope with temporal synthesizability. We use a binary label $l^{dt} \in \{0,1\}$ to indicate the spatial mode of DT sample $V$, {\em i.e.} `1' for SHDT, and `0' for TDT.

To suggest which EDTS method is best suited to synthesize $V$, an integer label $l^{md} \in \mathbb Z^+$ is used. We employed $M$ representative off-the-shelf EDTS methods as candidate synthesis methods, thus $l^{md} \in \{1,2,\dots,M\}$.
Then, we denote a DT sample $V$ with synthesizability score $s$ and the two associated labels $l^{dt}$ and $l^{md}$ as $(V, \mathcal Y)$, where
$\mathcal Y=\{l^{dt},s,l^{md}\}$ is the label space.
The task of learning dynamic texture synthesizability is to build a mapping $\mathcal F: V \to \mathcal Y=\{l^{dt},s,l^{md}\}$ from a training data set $\mathcal D= \{(V_i, \mathcal Y_i)\}_{i=1}^N$ to predict the label sets for unseen dynamic texture.
The problem of learning dynamic texture synthesizability can be formulated as
\begin{align}\label{eq:mapping}
\mathcal Y = \mathcal F(\mathbf F \circ V)
\end{align}
where $\mathbf F$ is a video feature extractor put on $V$.

The goal of learning dynamic texture synthesizability is to build the mapping $\mathcal F$. Then given a DT sample $V$, the task of predicting synthesizability is to estimate and assign the label sets $\mathcal Y=\{l^{dt},s,l^{md}\}$ for the sample $V$.

\subsection{Label space}
The three labels for annotation are not independent, but they are associated with each other.
Based on the spatial mode, dynamic textures can be divided into SHDTs and TDTs labeled as $l^{dt}$ in the training data, with or without synthesis capacity in space.
Accordingly, the annotation of synthesizability score $s$ for SHDT and TDT is different: SHDT with spatial and temporal synthesizability respectively, TDT only with temporal synthesizability.
In addition to the synthesizability score, the optimal synthesis method of each DT video in the training set was also recorded and labeled as $l^{md}$. Not all EDTS methods are able to synthesize both TDTs and SHDTs. Some methods can deal with the both, but the others have preference for one of them. Hence, $l^{md}$ should also be connected with $l^{dt}$.

To summarize, there are three type of labels for a DT: binary label for the spatial mode $l^{dt}$, continuous value to label synthesizability score $s$, discrete label for method index $l^{md}$. Obviously, this is a multi-label problem but more than that, because three labels are interrelated. Whether to compute spatial synthesizability or not is determined by label $l^{dt}$.
If $l^{dt}=0$ for TDT, we only compute temporal synthesizability score; whereas $l^{dt}=1$ for SHDT to compute both spatial and temporal aspects.

\subsection{Divide-and-conquer strategy}
To clarify this problem, given the dynamic texture training set $\mathcal D= \{(V_i, \mathcal Y_i)\}_{i=1}^N$ where every DT sample is associated with three labels, we want to learn a set of models to estimate the synthesizability of unknown test dynamic texture.
The mapping $\mathcal F: V \to \mathcal Y=\{l^{dt},s,l^{md}\}$ from one video to three lables is an one-to-many mapping, which is difficult to tackle.
To simplify the problem involved with three correlated labels, we break it down into three sub-tasks to solve the complex mapping $\mathcal F$: binary classification for SHDT and TDT, regression to predict synthesizability score, and an additional classier to suggest the ``best'' synthesis method.

\subsubsection{Learning binary label $l^{dt}$}
Binary classification of SHDT and TDT amounts to estimate the class posterior distribution $P_{dt}(l^{dt}|V)$ over the binary classes $l^{dt} \in \{0,1\}$, which can be given as
\begin{align}\label{eq:class_posterior}
l^{dt} = arg\max_{l^{dt}\in \{0,1\}} P_{dt}(l^{dt}|V)
\end{align}

\subsubsection{Learning synthesizability score $s$}
The synthesizability $s \in [0, 1]$ of a DT sample $V$ indicates the probability that EDTS methods will produce good synthesized results for $V$ before synthesizing it. We denote a DT sample $V$ with synthesizability score $s$ as $(V, s)$. The problem of learning dynamic texture synthesizability score can be formulated as
\begin{align}\label{eq:v_regress}
s = f(\mathbf F \circ V)
\end{align}
where $f$ is a function to learn synthesizability, e.g. regression model, and $\mathbf F$ is a video feature extractor.

\subsubsection{Suggest synthesis method index $l^{md}$}
We also suggest the ``best'' EDTS method by an additional classifier, which can be regarded as multi-class classification to assign a method label $l^{md}$ to indicate the optimal synthesis method for DT sample.
Since synthesis methods have their own abilities to deal with different spatial modes, we can introduce prior information indicated by $l^{dt}$ to constrain the possible methods for SHDTs or TDTs, which leads to a joint class posterior probability distribution:
\begin{align}
l^{md} = arg\max_{l^{md}} P_{md}(l^{md}|V, l^{dt})
\end{align}

The goal of learning dynamic texture synthesizability is to set up the function $f$, and build the class posterior probability distribution $P_{ld}$ and $P_{md}$. Then given a test DT sample $V$, the task of predicting synthesizability is to give the spatial mode label $l^{dt}$ and estimate synthesizability score value $s$, as well as suggest the optimal EDTS method label $l^{md}$ for the sample $V$.

\section{Method to Learn Dynamic Texture Synthesizability}\label{sec:method}
In this section, we firstly investigate the dynamic texture representation related to synthesizability. Next, the method of learning dynamic texture synthesizability is depicted in detail. Finally, how to use synthesizability to detect synthesizable regions in video is introduced.

\subsection{Dynamic texture representation}
For dynamic texture representation relevant to synthesizability, we revisit general spatial-temporal features and design a novel SCOP-DT descriptor for dynamic texture.
\subsubsection{LBP-TOP~\cite{zhao2007dynamic_pami}}
\emph{Local binary patterns from three orthogonal planes} (LBP-TOP)~\cite{zhao2007dynamic_pami} extended Local Binary Patterns (LBP)~\cite{ojala2002multiresolution} to 3D volume for dynamic texture analysis by calculating the LBP code in three orthogonal 2D cross profiles $XY$, $XT$ and $YT$ planes. LBP-TOP is invariant to local contrasts of dynamic textures, but can not depict the geometric aspects.

\subsubsection{C3D ConvNet}
In static texture recognition, Cimpoi {\em et al.}~\cite{cimpoi2014describing} have successfully ported object/scene description methods \emph{e.g.} convolutional neural networks (CNNs)~\cite{donahue2014decaf} to texture descriptors, and significantly outperform the state-of-the-art recognition rates established by specialized texture descriptors. Inspired by their work, here we use the generic video descriptors learned by CNNs to characterize DTs.
Recently, the architecture of deep 3-dimensional convolutional networks (C3D ConvNets)~\cite{tran2015learning} has been proposed to learn spatiotemporal features for video description. C3D trained the network with 3D filter kernels in space-time volume, which can be used as a generic video feature extractor, but not specific for dynamic textures.

\subsubsection{Shape-based co-occurrence patterns for dynamic textures}\label{subsec:scopdt}
In static texture analysis, Shape Co-occurrence Patterns (SCOPs)~\cite{Xia2017Texture,liu2014texture} proposed a kind of shape-based texture representation by using the co-occurrence patterns of shapes. Following the shape-based invariant texture analysis (SITA)~\cite{xia2010shape}, texture images are first represented by {\em tree of shapes} (the topographical map), where each shape is associated with some predefined geometrical and radiometric attributes.
SCOPs learned a set of co-occurrence patterns of shapes via clustering based on the hierarchical relationship of the shapes in the tree. Establishing co-occurrence patterns of shapes as codewords of dictionaries, a texture image can be encoded into a vector descriptor. SCOPs captured geometrical aspects of textures and high-order statistics between shape relationships, which demonstrated superior performance both on multiple texture image dataset and the complex scene image dataset.

The original SCOP is designed for static texture in 2D space. In this paper, we extend SCOP descriptor from image space to space-time dynamic texture by incorporating temporal cues implicitly. A natural way to extend the SCOP for DT is to treat a DT sequence as a 3D volume. This requires to extend the tree of shapes for 2D image to handle 3D space-time video, where each shape is 3D in volume rather than 2D in plane. This approach, while seemingly natural and sound, faces challenges such as dealing with varying frame rates or motion speed~\cite{ji2013wavelet}. In this case, to treat space and time equally in 3D cuboid domain may not be reasonable, owing to the different scale and occurrence of elements observed in space and motion. To bypass this problem, we propose an alternative method here, which implicitly captures the self-similarity behavior of DT sequence along time axis, referred to as SCOP-DT descriptor/feature in this paper.

\begin{figure*}[ht!]
  \centering
  \includegraphics[width=0.95\linewidth]{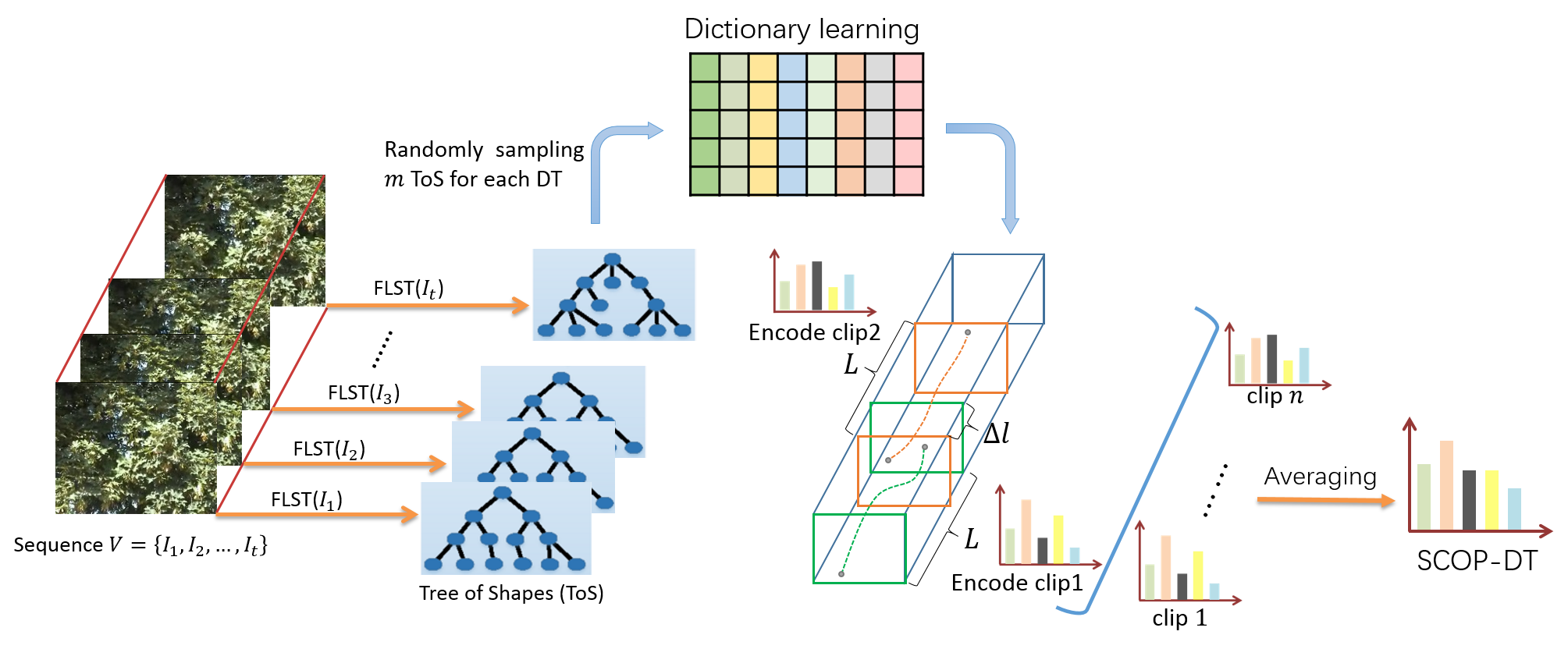}
\vspace{-3mm}
   \caption{The proposed representation of shape-based co-occurrence patterns for dynamic textures (SCOP-DT).}
   \label{fig:SCOP-DT}
\end{figure*}

The extraction of SCOP-DT in video is illustrated in Fig.~\ref{fig:SCOP-DT}. We use {\em fast level set transformation} (FLST)~\cite{monasse2000fast} to calculate {\em Tree of Shapes} (ToS) for each frame in DT video. Due to self-similarities between frame images in dynamic texture, the extracted ToS of each frame is also similar to each other. To construct dictionaries of shape co-occurrence patterns in ToSs, we randomly choose $m$ ToSs of $m$ frames from every training video to learn codewords in the dictionaries.
In the stage of encoding, based on the temporal consistency in DTs, we implicitly encode temporal information to incorporate both shape and dynamic aspects for joint spatial-temporal representation.

More precisely, suppose a DT sample $V$ is a sequence of $T$ images $V = \{I_0,\cdots, I_t, \cdots, I_{T-1}\}$ along $t$ time axis. The {\em Tree of Shapes} (ToS) of the $t$-th frame $I_t$ is calculated by {\em fast level set transformation} (FLST): $ToS_t = FLST(I_t)$. Then we randomly choose $m$ ToSs of $m$ frames from each video in the training sets to learn the dictionary $Dict$, e.g. via clustering.
In the coding step, we slice a video into a set of $L$ frame clips with stride of $\vartriangle\!l$ frame interval between two consecutive clips. To include temporal dependencies in SCOP-DT, we project the ToSs of a clip with consecutive $L$ frames onto the dictionary $Dict$ and compute a coding vector. The coding vector $SCOP\!-\!DT_{clip_{0,L-1}}$ of a $clip_{0,L-1}$ with a set of $ToS_{0,L-1} = \{ToS_0, ToS_1,\cdots, ToS_{L-1}\}$ can be encoded by:
\begin{align}
SCOP\!-\!DT_{clip_{0,L-1}} = Encode(ToS_{0,L-1},\; Dict)
\end{align}
where $Encode(\cdot,\cdot)$ denote projecting the set of $ToS$s onto the learned $Dict$ to calculate the coding vector.

We sample ToS sequences set $\{ToS_{n\vartriangle l,n\vartriangle l+L-1}\}_{n=0}^N$ with stride of $\vartriangle\!l$ sliding window along $t$ time axis for all clips in a video, with maximum $N\vartriangle\!l+L-1\leq T$. The encoding of $SCOP\!-\!DT_{clip_{n\vartriangle l,n\vartriangle l+L-1}}$ can be denoted by
\begin{align}
SCOP\!-\!DT_{clip_{n\vartriangle l,n\vartriangle l+L-1}} = Encode(ToS_{clip_{n\vartriangle l,n\vartriangle l+L-1}},Dict)
\end{align}

To extract SCOP-DT feature, the coding vectors of all clips in $V$ are averaged to form a final SCOP-DT video descriptor.
\begin{align}
SCOP\!-\!DT(V) = \frac 1{N+1} \sum_{n=0}^N SCOP\!-\!DT_{clip_{n\vartriangle l,n\vartriangle l+L-1}}(V)
\end{align}

It is noticed that this approach implicitly incorporate time information by simultaneously encoding $L$ frames into one coding vector, which implies self-similarities and temporal stationarity in DTs. To further improve the stability of the descriptor, the mean of coding vectors over all clips is used. By taking the average response over time, it effectively suppresses the variations of the computed encoding vectors of all clips.

\subsection{Framework of learning synthesizability}

\begin{figure*}[ht!]
  \centering
  \includegraphics[width= 0.9 \linewidth]{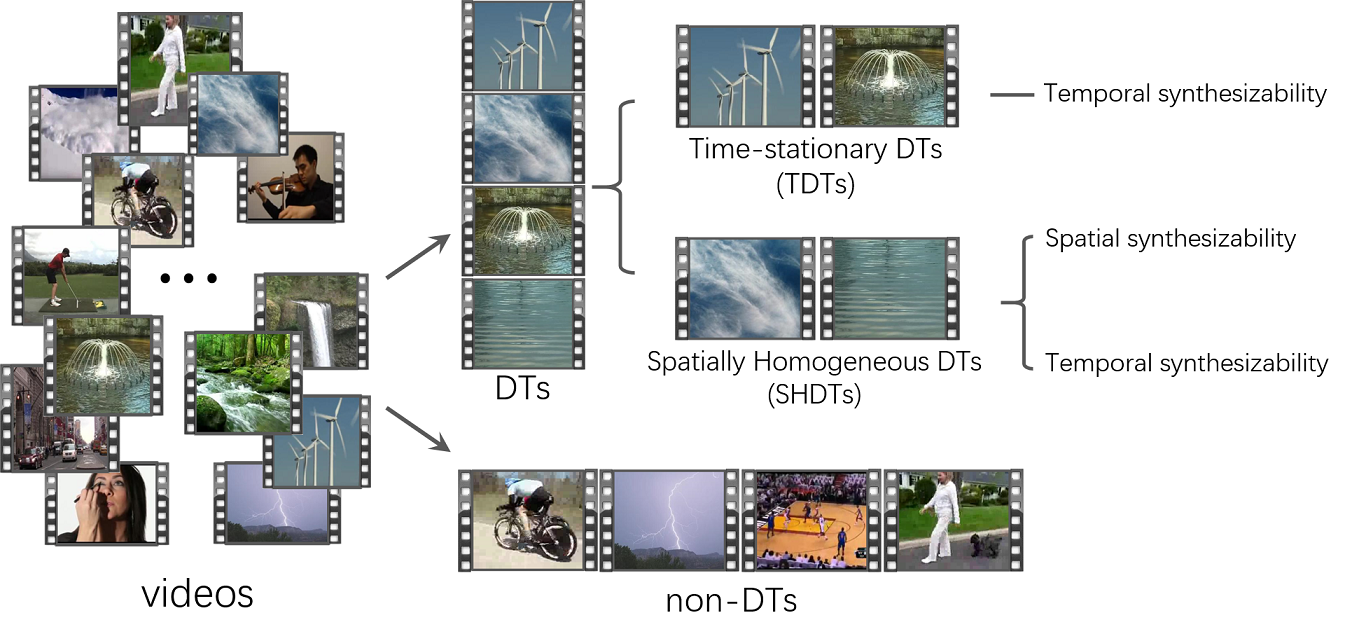}
\vspace{-3mm}
   \caption{The hierarchical learning scheme of automatic selection of dynamic textures, separation of SHDTs and TDTs, and prediction of spatial and temporal synthesizability.}
   \label{fig:hierarchy}
\end{figure*}

Having described dynamic texture representation relevant to synthesizability, we now present the technical framework of learning processing. Our ultimate goal is to predict synthesizability of dynamic texture sample, but we intend to generalize this problem into more unconstrained setting. We would like to first retrieve dynamic textures from general video resources (including DT and non-DT), and then pre-compute synthesizability for SHDT and TDT according to DT's spatial mode.
To this end, as illustrated in Fig.~\ref{fig:hierarchy}, we proposed a hierarchical architecture to deal with the routine from generic video sequences to dynamic textures, further distinguish DT into SHDT and TDT, and output synthesizability score at last.
This leads to a 2-level partition. We retrieve DTs as positive examples to separate DTs from non-DT videos in the 1st-level. The partition for SHDTs and TDTs in the 2nd-level is performed by binary classification. The prediction of synthesizability in the final stage is conducted as a regression problem. An additional classifier is used to suggest the ``best'' EDTS method.
We start with dynamic texture retrieval, to move on to binary classification for SHDTs and TDTs, and to prediction of synthesizability as well as suggestion on suitable EDTS methods.

The overall step-by-step procedure in Fig.~\ref{fig:hierarchy} contains three main stages as follows:
\begin{itemize}
\item[-] retrieve dynamic textures to separate DTs from non-DT videos;
\item[-] binary classification to distinguish DTs into SHDTs and TDTs;
\item[-] predict synthesizability scores by regression and suggest suitable EDTS methods by classification.
\end{itemize}

\subsubsection{Retrieve DTs from videos}
\begin{figure}[ht!]
  \centering
  \includegraphics[width= 0.7 \linewidth]{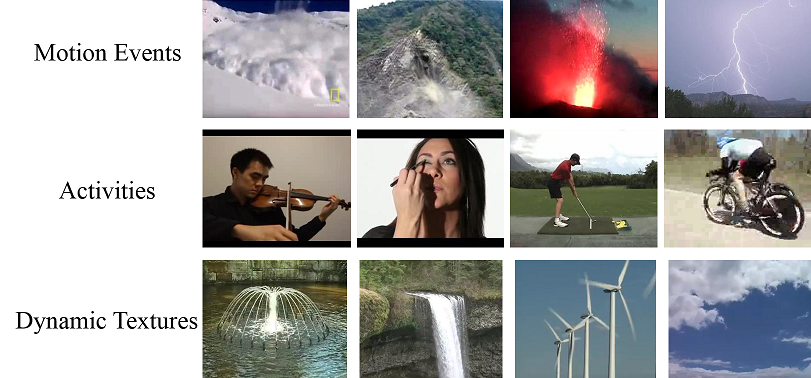}
\vspace{-3mm}
   \caption{Three kinds of visual motions: motion events, activities and DTs.}
   \label{fig:DTEventsActs}
\end{figure}

In nowadays, with easy access to portable sensor devices (e.g. mobile phones), videos being shared have become ubiquitous on the Internet.
The explosive growth of video data makes video processing popular and imperative in recent years. As for dynamic textures, on one hand, big data provides a great diversity of video source, on the other hand, it turns out to be a huge workload to pick out dynamic texture exemplars from video data.
Polana and Nelson categorized visual motion recorded as video into three kinds~\cite{polana1997temporal}: activities, motion events, and DTs. We give several examples for each category in Fig.~\ref{fig:DTEventsActs}. Activities and motion events are more difficult to synthesize than dynamic textures. In this paper, we focus on dynamic textures and want to retrieve DTs in a variety of videos when dynamic textures, motion events and activities all occur.

We train a classifier to distinguish dynamic textures from activities and dynamic scenes (mostly motion events). The dynamic texture dataset we collect delivered the positive samples (totally 1729 DTs). Maryland~\cite{shroff2010moving} and YUPENN~\cite{derpanis2012dynamic} dynamic scene datasets, and the UCF101 action dataset~\cite{soomro2012ucf101} as the negative ones (events/activities), consist of 13,520 videos in all. Generally speaking, DTs are only a small fraction of videos. Hence the number of negative samples (non-DTs) is almost an order of magnitude than the positive ones (DTs) in retrieval. We choose C3D descriptor for representation in the retrieval task because it is a generic video descriptor confirmed by many video recognition and classification tasks. The universality of C3D makes it equally effective to characterize DTs and non-DTs without much specialization, considering that all kinds of videos are included in retrieval. Random Forest was used as the regression model with C3D descriptor as feature. The regression score is taken as DT ``textureness'' to quantify the chance that a video is a dynamic texture.

\subsubsection{Binary classification for SHDTs and TDTs}
After picking out DTs, we should split up DTs into SHDTs and TDTs straightforward to accommodate the joint spatial-temporal synthesis or only temporal synthesis. We train a classifier to distinguish DTs into SHDTs and TDTs by using the proposed SCOP-DT feature. As SCOP-DT is able to characterize the geometrical and radiometric attributes of textures inherited from SCOP~\cite{Xia2017Texture}, this property can be useful to distinguish SHDTs and TDTs due to their difference mainly in structural aspects of appearance. To tackle the binary classification of SHDTs and TDTs, we use the proposed SCOP-DT as feature to train a binary-class SVM classifier.
Considering that early researches in image recognition have shown that combining multiple descriptors is very useful to improve classification performance~\cite{grauman2005pyramid,lazebnik2006beyond}, SCOP-DT has been further combined with C3D and LBP-TOP in ensemble SVMs~\cite{yang2016dynamic} for binary classification of SHDTs and TDTs.

\subsubsection{Learning the synthesizability of dynamic texture samples}
Now that DTs are classified into two subsets: SHDTs and TDTs, the last step is to predict synthesizability for them.
SHDT samples have spatial and temporal synthesizability, while TDTs merely cope with temporal synthesizability.

\paragraph{Learning synthesizability:}
We suppose that synthesizability score is learnable and predictable, which can be formulated as a regression problem.
Given the DT samples and labeled synthesizability scores $\{(V_i, s_i)\}_{i=1}^N$ in the training set, we formalise the learning problem as a regression model $f$:
\begin{align}\label{eq:vi_regress}
s_i = f(\mathbf F \circ V_i)
\end{align}

Let the DT sample $V$ be described by a set of $K$ features ${\{\mathbf x^k\}}_{k=1}^K$,
\begin{align}\label{eq:feature}
\forall 1 \leq k \leq K, \,\, \mathbf x^k = \mathbf F_k \circ V,
\end{align}
where $\mathbf F_k$ is feature extractor e.g. SCOP-DT, C3D and LBP-TOP.

By the feature representation of DT sample $V_i$ in~\eqref{eq:feature}, we denote the regression model as
\begin{align}\label{eq:x_regnonlinear}
f_k(\mathbf x_i^k) = \mathbf w_k^T \phi(\mathbf x_i^k)+b_k,\quad f_k(\mathbf x_i^k) = s_i,
\end{align}
where $\mathbf w_k$ and $b_k$ are parameters of regression model $f_k$ corresponding to feature representation $\mathbf F_k$, with $^T$ denoting the vector transpose, and $\phi$ can be linear or nonlinear mapping.

To build the regression model~\eqref{eq:x_regnonlinear}, we train the model on training set $\big \{\big ({\{\mathbf x_i^k\}}_{k=1}^K, s_i \big )\big \}_{i=1}^N$ to determine parameters $\mathbf w_k$ and $b_k$ by
\begin{align}\label{eq:lossfunc}
(\mathbf w_k,b_k) = arg\min_{(\mathbf w_k,b_k)}\sum_{i=1}^N(f_k(\mathbf x_i^k) - s_i)^2.
\end{align}

Then for the test DT sample $V_{test}$ with unknown synthesizability score, we can use the trained regression model~\eqref{eq:x_regnonlinear} to predict synthesizability score $s_{test}^k$ with feature representation $\mathbf F_k$.
The predicted synthesizability score $s_{test} \in [0,1]$ is a computable index to quantify how well a dynamic texture can be synthesized by only analysing the original sample, where the bigger score the better synthesizability.

\paragraph{Predicting synthesizability:}
In the spirit of Dai {\em et al.}'s~\cite{dai2014synthesizability} approach in utilization of multi-feature combination to predict static texture synthesizability, we follow their policy for dynamic texture synthesizability prediction. But unlike Dai {\em et al.}'s method by simply concatenating multiple features for combination, we aggregate different features on decision level for more robust fusion.
In general, it is beneficial to combine multiple descriptors in machine learning~\cite{grauman2005pyramid,lazebnik2006beyond}. The naive solution to combination is to concatenate different descriptors into one vector, yet faced with some deficits. On one hand, a possible problem of creating a large input vector for a machine learning classifier is that the input vector becomes of very large dimensionality, which may lead to overfitting and degenerate generalization performance~\cite{abdullah2009spatial}. On the other hand, different features lying in different feature space with different scales are likely incompatible in a concatenation.

We combine multiple features to predict dynamic texture synthesizability by aggregating different features on decision level. We use three features SCOP-DT, C3D and LBP-TOP in combination.
Recently, ensemble of classifiers have been used to combine features efficiently in dynamic texture recognition~\cite{yang2016dynamic}. Therefore, we resort to feature combination on decision level by ensemble schedule. There are two merits as for fusion on decision level. We don't need to consider the normalization problem of concatenating different descriptors directly. Besides, the choice of classifiers can be more flexible for different type of descriptors, considering that every feature may be in favour of a specific kind of classifier.

We select the optimal regression model relevant to each feature, e.g. Support Vector Machine (SVM) or Random Forest (RF). Regression models on the training data with labeled synthesizability scores are trained for each feature respectively as depicted in formulation~\eqref{eq:x_regnonlinear}. The trained models are then used to predict the synthesizability of a given video. Thus for a video, $K$ synthesizability scores can be predicted with respect to $K$ features. Then $K$ scores are weighted averaged to form a final synthesizability prediction score. The weights are set manually according to the performance of each individual feature for effective combination. The final output prediction score $s$ is a weighted average of $s^k$ given the $K$ different features of the unknown test sample $V_{test}$:
\begin{equation}\label{eq:weight_ave}
s(V_{test}) = \sum_{k=1}^K \alpha_k f_k(\mathbf F_k \circ V_{test}),
\end{equation}
where $\alpha_k$ are weighting factors \emph{w.r.t.} $K$ features C3D, LBP-TOP and SCOP-DT.

\begin{figure*}[htb!]
  \centering
  \includegraphics[width= 0.9 \linewidth]{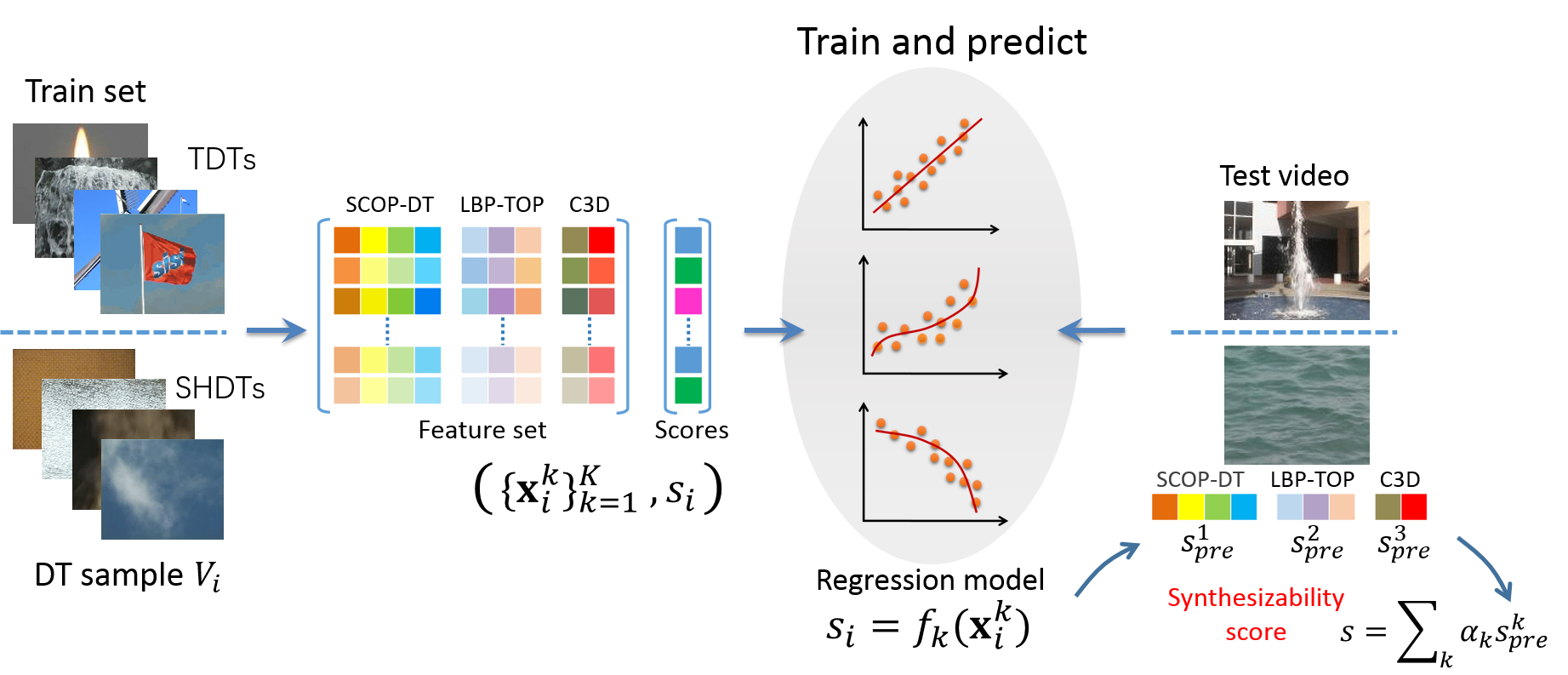}
\vspace{-3mm}
   \caption{Learning and predicting synthesizability by aggregating features to train regression models.}
   \label{fig:aggregate_feature}
\vspace{-3mm}
\end{figure*}

The scheme of learning and predicting synthesizability by aggregating features is illustrated in Fig.~\ref{fig:aggregate_feature}. Firstly, we use regression models SVM and RF for every single feature to predict synthesizability, and compare the performance of two regression models. Secondly, we choose the optimal feature and regression model among them, and set weights manually for feature combination in formula~\eqref{eq:weight_ave}. Finally, the predicted synthesizability scores of feature combination on decision level are output.

\subsubsection{Suggest synthesized methods}
We not only predict synthesizability score for a given DT sample, but also suggest the ``best'' EDTS method to synthesize it. The optimal synthesized method of each DT in the training set is also recorded as a method label $l^{md}$ along with the synthesizability score. We can recommend the ``best'' EDTS method by an additional classifier. We combine C3D and SCOP-DT features in Ensemble SVMs~\cite{yang2016dynamic} for classification. Algorithm~\ref{alg1} presents the implementation pipeline of predicting synthesizability and suggesting optimal EDTS method in our approach.

{
    \begin{algorithm}[htb!]
    \caption{Learning dynamic texture synthesizability}
    \begin{algorithmic}[1]\label{alg1}
    \REQUIRE ~~\\
    the training DT set $\mathcal D= \big\{(V_i, \{s_i,l_i^{md}\})\big\}_{i=1}^N$ with labeled synthesizability score $s$ and labeled optimal synthesized method $l^{md}$; test DT sample $V_{test}$ with unknown synthesizability.
    \ENSURE ~\\
    the predicted synthesizability score $s_{test}$ and the suggested synthesized method $l_{test}^{md}$ for $V_{test}$.
    \STATE calculate all features C3D, SCOP-DT and LBP-TOP for the training set $\mathcal D$ and test DT sample $V_{test}$.
    \FOR {$i=1,\cdots, N$}
    \FOR {$k=1,\cdots, K$}
    \STATE calculate feature vector $\mathbf x^k_i = \mathbf F_k \circ V_i$.
    \ENDFOR
    \ENDFOR
    \STATE learn the regression models $\{f_1,\cdots,f_K\}$ with $\big \{\big ({\{\mathbf x_i^k\}}_{k=1}^K, s_i \big )\big \}_{i=1}^N$ by Eq.~\eqref{eq:lossfunc}.
    \STATE train classifier Ensemble SVMs with $\big \{\big ({\{\mathbf x_i^k\}}_{k=1}^K, l_i^{md} \big )\big \}_{i=1}^N$.\label{code:classify}
    \FOR {$k=1,\cdots, K$}
    \STATE calculate feature vector $\mathbf x^k_{test} = \mathbf F_k \circ V_{test}$;
    \STATE compute the predicted synthesizability score $s^k_{test}$ using $f_k$ and $\mathbf x^k_{test}$ by Eq.~\eqref{eq:x_regnonlinear}.
    \ENDFOR
    \STATE combine $\{s^1_{test},\cdots,s^K_{test}\}$ via Eq.~\eqref{eq:weight_ave} to predict the final synthesizability score $s_{test}$ for $V_{test}$.
    \STATE suggest the optimal synthesized method $l^{md}_{test}$ for $V_{test}$ with the trained Ensemble SVMs in Step~\ref{code:classify}.
    \end{algorithmic}
    \end{algorithm}
    \par
}

\subsection{Detecting synthesizable regions}
In natural world, dynamic textures normally appear as visual phenomena with cluttered background in complex scene. Correspondingly, DTs captured in unconditional circumstances only occupy parts of videos with uncertain and irregular shapes, which are readily inappropriate to use the whole video as input exemplar for DT synthesis. Therefore, it would be beneficial to tailor videos into regions with good synthesizability by discarding undesirable background. To this end, we use the dynamic texture detection method~\cite{fazekas2009dynamic} to detect the rough and irregular DT regions in video at first. Then, the detected region is trimmed into regular shape aimed at good synthesizability.
100 rectangular subregions were randomly sampled within the detected region to compute and compare their synthesizability scores. The most synthesizable subregion is then suggested as shown in Fig.~\ref{fig:detection_trimming}. The subregions are spatially homogeneous dynamic textures (SHDTs) and can be synthesized in spatial extent. 
\section{Data Collection and Annotation}\label{sec:dataset}
Since we want to formulate the prediction of synthesizability as a regression problem, it is necessary to compile a collection of annotated data in terms of synthesizability.
There are several established benchmark datasets from dynamic texture community, which have been mainly targeted on classification issues yet. In order to fit into our problem, we compiled a dynamic texture dataset and manually annotated it with synthesizability score. Most of the dynamic texture examples came from available existing datasets e.g. UCLA~\cite{saisan2001dynamic}, DynTex~\cite{peteri2010dyntex} and Spacetime Texture Dataset~\cite{derpanis2012spacetime}. To enlarge the dataset size, we selected some samples belonging to dynamic texture category from two dynamic scene datasets: Maryland ``In-The-Wild''~\cite{shroff2010moving} dataset and the YUPENN Dynamic Scenes~\cite{derpanis2012dynamic} dataset. In comparison to dynamic textures, dynamic scenes are composed of moving scene elements with certain spatial layout, where several inner-related regions of different dynamic patterns appear in complex settings, (e.g. burning fire with billowing smoke in forest, and a downtown street scene composed of the pedestrians, vehicular traffic and flashing lights). Thus, dynamic textures can be viewed as a kind of particular dynamic scenes in simpler settings, typically with the field of view restricted to a single uniform dynamic region. Therefore we also picked out some dynamic texture samples from the dynamic scene dataset. Finally, we ended up with a dataset of 1729 DT samples, among which there are 452 SHDTs and 1277 TDTs respectively.

Since none of the EDTS methods can always perform well on all kinds of dynamic textures, several representative state-of-the-art methods were tested, in order to cover different aspects of EDTS methods and provide a comparison across these synthesis results. The selected algorithms comply with the following principles: the classical LDS model~\cite{doretto2003dynamic}, two compact representation models~\cite{xia2014synthesizing} SN-textons and AR-textons, copy-based method Graphcut Textures~\cite{kwatra2003graphcut}, two CNN-based methods STGConvNet~\cite{xie2016synthesizing} and Yang {\em et al.}'s work~\cite{yang2016stationary}, the latter namely Gatys-DT here, which follows Gatys {\em et al.}'s static texture synthesis method~\cite{gatys2015texture}.
The six dynamic texture synthesis methods have their own preference for two types of dynamic textures SHDTs and TDTs. All six methods were used for SHDTs; whereas for TDTs, only three methods LDS model, Graphcut Textures and STGConvNet were included, because the other three methods work on synthesis in joint spatiotemporal domain, which are unable to customize temporal synthesis specifically.

\begin{figure}[htb!]
  \centering
  \includegraphics[width= 0.99\linewidth]{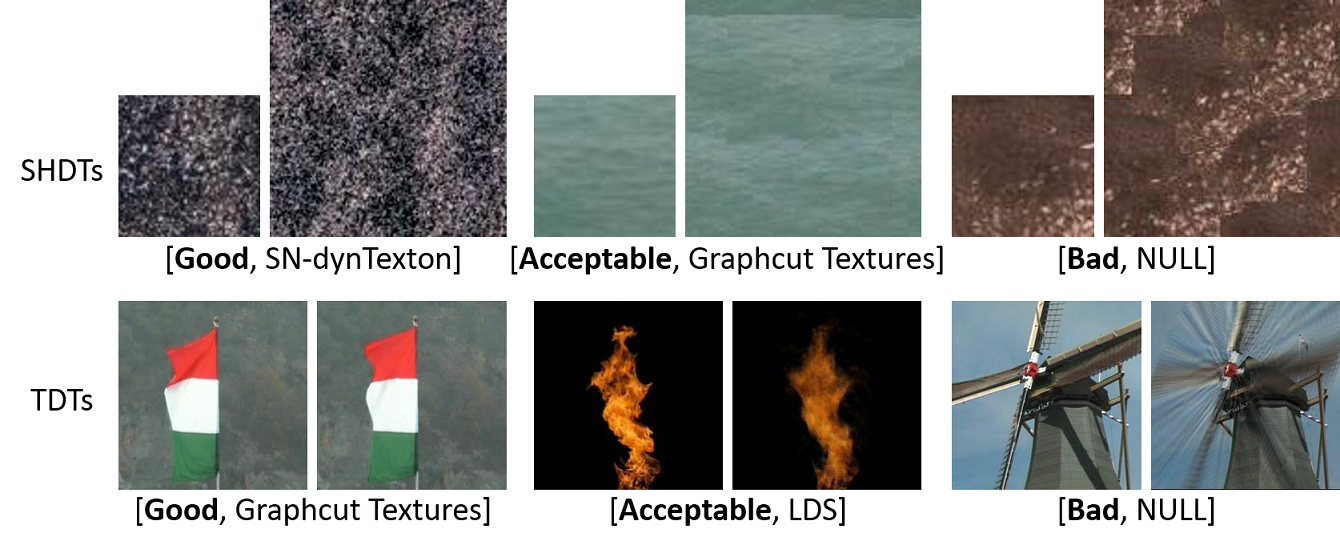}
\vspace{-3mm}
   \caption{Some examples from our dataset with their annotations of synthesizability. In each pair, left: DT sample; right: synthesized result.}
   \label{fig:database}
\vspace{-3mm}
\end{figure}

We synthesize each dynamic texture example with corresponding available algorithms. Several synthesis results are compared against each other. Then we manually annotated the synthesizability score of each dynamic texture sample as the ``goodness'' level of the best synthesized result. The synthesizability score of a dynamic texture sample is annotated as follows: given a sample, the annotator got all the synthesis results and chose the best one to score. Following the work of~\cite{dai2014synthesizability}, the goodness of the synthesizability was divided into 3 levels: good, acceptable, and bad, with the quantitative scores assigned as 1, 0.5 and 0 respectively. The best synthesis method of each DT example was also recorded for ``good'' and ``acceptable'' DTs; ``bad'' ones were assigned to ``NULL''.
The synthesis results can only be evaluated qualitatively by observing the perceptual quality. An ideal synthesized dynamic texture should look perceptually similar to the input example when perceived by a human observer, and should not have visible artifacts such as seams, blocks or corrupt elements, should not show discontinuity or missing frames during playback. Since the synthesized video should be as natural as possible while maintaining equivalent visual perception to the input, the verbatim copying reproduction of salient repeated parts is undesired, if not the case in the original. The final outcome of synthesizability score annotated for a DT example is the ``goodness'' of the synthesized result that an expert annotator considered best. See Fig.~\ref{fig:database} for examples of such annotation.
TDTs can only be synthesized along time, whereas synthesis for SHDTs is practicable both in space and time. Therefore, we annotated TDTs with temporal synthesizability score, and SHDTs are labeled with spatial and temporal synthesizability score respectively. For temporal synthesizability of 1277 TDTs, 25.69\% samples were labeled bad, 24.67\% acceptable and 49.65\% good. For 452 SHDTs, spatial and temporal synthesizatility scores were both annotated: spatially 29.42\% bad, 36.95\% acceptable and 33.63\% good; temporally 4.65\% bad, 32.30\% acceptable and 63.05\% good.
 
\section{Results and Validation}
\label{sec:experiment}

In this section, we evaluate all sub-tasks of the hierarchical procedure in Figure~\ref{fig:hierarchy}. We run through the experiments step-by-step to evaluate each task on the dataset. Both quantitative and qualitative experiments are reported. All the results are available at \url{http://captain.whu.edu.cn/project/DTsynthesizability.html}, where one can checked the videos.

\subsection{Quantitative evaluation}
$50\%$ of the dataset videos were used for training, the rest for testing. We report results over 100 random training-testing splits in all quantitative experiments.

\subsubsection{Retrieve DTs from videos}
\begin{figure}[htb!]
\vspace{-3mm}
  \centering
  \includegraphics[width= 0.6 \linewidth]{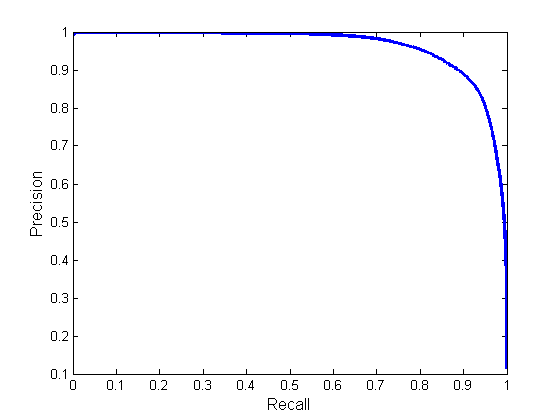}
\vspace{-3mm}
   \caption{The precision-recall curve for DT retrieval in videos.}
   \label{fig:PRcurve_6setTexSceneUCF}
\vspace{-2mm}
\end{figure}

We retrieve DTs in videos using Random Forest with C3D feature. There are 15,249 videos, of which 1729 videos are DTs. We evaluated the precision for different levels of recall in the retrieval task. The precision-recall curve is plotted in Figure~\ref{fig:PRcurve_6setTexSceneUCF}. The average precision is $96.23\%$ when half for training.

\subsubsection{Binary classification for SHDTs and TDTs}
There are $452$ SHDTs and $1277$ TDTs in our synthesizability dataset. We use SCOP-DT feature to distinguish DTs into SHDTs and TDTs. To verify the proposed SCOP-DT for DTs, we compare it to the approach in~\cite{yang2016dynamic}, which randomly select several frames in a DT to extract SCOPs for each selected frame separately and perform dynamic texture recognition in the late fusion ensemble architecture. We set $L=16$ for SCOP-DT. In accord with this, we randomly choose $16$ frames for the approach in ~\cite{yang2016dynamic}. The comparison is shown in Table~\ref{tab:scop_compare}, where the SCOP-DT we propose for DTs outperformed the usage of SCOP in~\cite{yang2016dynamic} by a large margin. The results confirmed that SCOP-DT containing implicit temporal information can benefit the description of DTs.

\begin{table}[htb!]
\centering
\caption{Comparison of the proposed SCOP-DT and randomly choosing frames to extract SCOP in~\cite{yang2016dynamic}.}
\label{tab:scop_compare}
\begin{tabular}{p{7em} |c |c  }
\hline
Features
\centering       &SCOP-random~\cite{yang2016dynamic}     &SCOP-DT     \\
\hline
Accuracy (\%)
\centering       &84.56     &92.65          \\
\hline
\end{tabular}
\end{table}

Considering that early researches in image recognition have shown that combining multiple descriptors is very useful to improve classification performance~\cite{grauman2005pyramid,lazebnik2006beyond}, SCOP-DT was combined with C3D and LBP-TOP in ensemble SVMs~\cite{yang2016dynamic} for binary classification of SHDTs and TDTs.
We evaluated all 3 single features and their combination in Table~\ref{tab:biclassify}. The accuracy of SCOP-DT is superior to both LBP-TOP and C3D. As expected, the feature combination further promotes classification rate, which gets 96.06\%.

\begin{table}[htb!]
\small
\centering
\caption{Binary classification for SHDTs and TDTs using LBP-TOP, C3D, SCOP-DT features and their combination.}
\label{tab:biclassify}
\begin{tabular}{p{7em} |c |c |c |c }
\hline
Features
\centering       &LBP-TOP    &C3D   &SCOP-DT    &\textbf{Combined} \\
\hline
Accuracy (\%)
\centering       &89.49   &91.56  &92.65  &\textbf{96.06}          \\
\hline
\end{tabular}
\end{table}

\subsubsection{Prediction of synthesizability}
For prediction of synthesizability, all 3 single features C3D, LBP-TOP, SCOP-DT and their combination were evaluated for SHDTs and TDTs respectively.
For quantitative evaluation, we performed two-level retrieval tasks and evaluated the average precision of synthesizability prediction with individual feature and multiple features: (1) retrieve videos with ``good'' scores ($\geq$good); (2) retrieve videos with ``good'' or ``acceptable'' scores ($\geq$acceptable).

\paragraph{Results on SHDTs:}
In prediction of spatial synthesizability for SHDTs, we also compared LBP-TOP, C3D, and SCOP-DT to the features used by Dai {\em et al.}~\cite{dai2014synthesizability} for static texture synthesizability. To utilize the static features in DTs, we randomly select 16 frames in each DT to extract features in~\cite{dai2014synthesizability} for each frame and compute the average feature over 16 frames.
The retrieval experiments of $452$ SHDTs are shown in Table~\ref{tab:synthesizability_SHDT_feature}. C3D with random forest and SCOP-DT with SVM have better performance than LBP-TOP and Dai {\em et al.}~\cite{dai2014synthesizability}. What's more, for spatial synthesizability prediction, SCOP-DT got the average precision score with 66.63\% for $\geq$good, and had an advantage over others by 6\% at least. Then we use the two fine features C3D and SCOP-DT for combination with the weighting factors set 0.5 for both, which can improve the average precision a little.

For $452$ SHDTs, from the precision scores (spatial synthesizability $\geq$acceptable 92.65\%, and $\geq$good 67.43\%; temporal synthesizability $\geq$acceptable 98.89\%, and $\geq$good 88.26\%), we can conclude that dynamic texture synthesizability is learnable and predictable. The table shows that spatial synthesizability is harder to predict than temporal synthesizability, because DTs usually exhibit strong self-correlation in time and make it relatively easy to synthesize new samples only along time. However, the degree of repetition and homogeneity is much lower in space than time. Also, the complex structures in 2D space make it more difficult to synthesize spatially, whereas dynamics in time is limited to 1D direction with its simplicity to some extent.

\begin{table*}[htb!]
\scriptsize
\centering
\caption{The average precision (\%) of synthesizability prediction for SHDTs with individual features and their combination}
\label{tab:synthesizability_SHDT_feature}
\begin{tabular}{c |c |c |c |c |c |c |c |c |c |c} 
\hline
\multicolumn{2}{c|}{Features}& \multicolumn{2}{c|}{LBP-TOP}& \multicolumn{2}{c|}{C3D}& \multicolumn{2}{c|}{SCOP-DT} &{C3D\,+\,SCOP-DT} &\multicolumn{2}{c}{Dai {\em et al.}~\cite{dai2014synthesizability}} \\
\hline
\multicolumn{2}{c|}{Regression methods}   &RF   &SVM  &RF  &SVM &RF &SVM   &RF + SVM &RF &SVM  \\
\hline
\multirow{2}{*}{spatial synthesizability}& $\geq$acceptable &77.85 &77.41 &91.47 &88.56 &85.80 &91.17 &\textbf{92.65} &90.77 &78.23\\
\cline{2-11}
& $\geq$good &40.17 &39.19	&60.45	&55.07	&51.17	&66.63 &\textbf{67.43} &58.64	&41.27\\
\hline
\multirow{2}{*}{temporal synthesizability}& $\geq$acceptable &97.60	&97.33	&98.84	&98.25	&97.47	&98.56 &\textbf{98.89} &- &-\\
\cline{2-11}
& $\geq$good &73.04	&70.51	&85.72	&82.79	&72.61	&85.93 &\textbf{88.26} &-	&-\\
\hline
\end{tabular}
\end{table*}

\paragraph{Results on TDTs:}
The total number of TDTs is $1277$ and the retrieval experiments are shown in Table~\ref{tab:synthesizability_TDT_feature}. For individual features, LBP-TOP with random forest got the best retrieval accuracy, and the performance of three features is more or less comparative. Thus, we combine all three features by equal weight of $\frac 1{3}$ to predict temporal synthesizability, by which we get the highest precision scores (96.20\% for $\geq$acceptable, and 90.30\% for $\geq$good). As already pointed out, the table also suggests that temporal synthesizability is not too difficult to predict, in line with simpler synthesis in time than space. The retrieval experiments confirm that we can expect a very high precision when a fraction of well-synthesizable TDT examples need to be retrieved. It is very useful to choose synthesizable DTs from internet videos in unconstrained circumstances.

\begin{table*}[htb!]
\scriptsize
\centering
\caption{The average precision (\%) of synthesizability prediction for TDTs with individual features and their combination}
\label{tab:synthesizability_TDT_feature}
\begin{tabular}{c |c |c |c |c |c |c |c |c |c} 
\hline
\multicolumn{2}{c|}{Features}& \multicolumn{2}{c|}{LBP-TOP}& \multicolumn{2}{c|}{C3D}& \multicolumn{2}{c|}{SCOP-DT} &{LBP-TOP\,+\,C3D\,+\,SCOP-DT} &{C3D\,+\,SCOP-DT} \\
\hline
\multicolumn{2}{c|}{Regression methods}   &RF   &SVM  &RF  &SVM &RF &SVM   &RF + RF + SVM  &RF + SVM\\
\hline
\multirow{2}{*}{\makecell{temporal\\ synthesizability}}& $\geq$acceptable &94.23	&93.15 &92.59 &91.18 &84.27	&93.50 &\textbf{96.25} &94.84\\
\cline{2-10}
& $\geq$good &86.89	&81.25 &81.78 &76.83 &66.83 &83.09 &\textbf{90.43} 	&87.26\\
\hline
\end{tabular}
\end{table*}

\subsubsection{Suggest synthesized methods}
For a given DT, we use an additional classifier to suggest the ``best'' EDTS method to synthesize it. For spatial-temporal synthesis of SHDTs, there are 4 synthesized methods: SN-textons, AR-textons, Graphcut Textures and Gatys-DT. Three choices of synthesized methods: LDS, Graphcut Textures and STGConvNet are used for temporal synthesis of TDTs. For simplicity, we choose two effective features C3D and SCOP-DT along with their combination for classification and the classification results are shown in Table~\ref{tab:classification_suggest_method}.

\begin{table}[ht!]
\small
\centering
\caption{Classification results of suggesting the `best' EDTS method to synthesize DTs.}
\label{tab:classification_suggest_method}
\begin{tabular}{p{7em} |c |c |c }
\hline
Features
\centering       &C3D   &SCOP-DT   &\textbf{Combined} \\
\hline
For SHDTs (\%)
\centering       &70.80   &73.26   &\textbf{75.76}    \\
\hline
For TDTs (\%)
\centering       &80.86   &80.88   &\textbf{82.39}    \\
\hline
\end{tabular}
\vspace{-3mm}
\end{table}

\subsection{Qualitative evaluation}
\begin{figure*}[ht!]
\vspace{-3mm}
 \centering{
    \subfloat[Synthesizability score: 0.81;]{
     \label{subfig:571b310}
    \includegraphics[height = 0.07\linewidth]{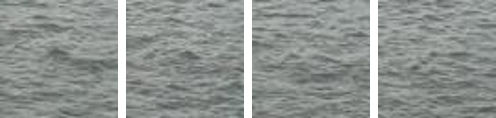}
 }
   \hspace{2mm}
    \subfloat[Synthesis method: Graphcut Textures; Annotated score: 1]{
     \label{subfig:571b310_GC_s2t15}
    \includegraphics[height = 0.14\linewidth]{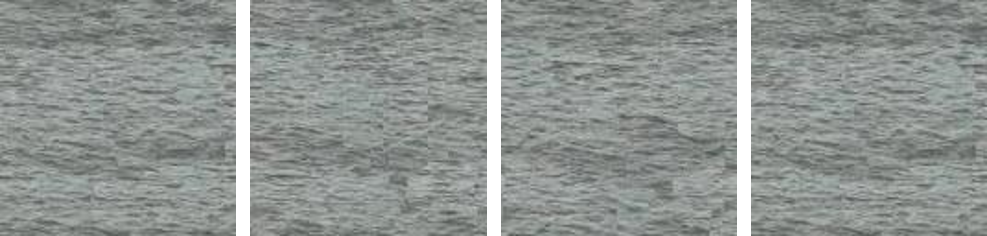}
 }
 \vspace{-3mm}

    \subfloat[Synthesizability score: 0.44;]{
     \label{subfig:55fa110}
    \includegraphics[height = 0.07\linewidth]{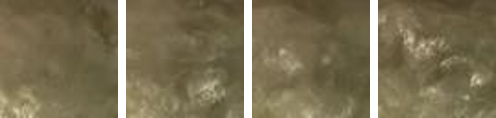}
 }
   \hspace{2mm}
    \subfloat[Synthesis method: Gatys-DT; Annotated score: 0.5]{
     \label{subfig:55fa110_Gatys_nfr50_st1_5}
    \includegraphics[height = 0.14\linewidth]{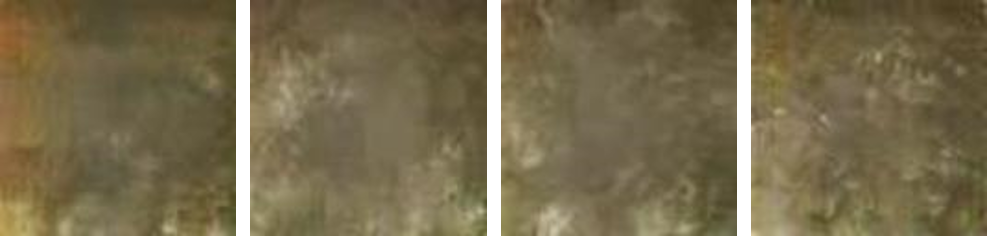}
 }
 \vspace{-3mm}

    \subfloat[Synthesizability score: 0.10;]{
     \label{subfig:6450810}
    \includegraphics[height = 0.07\linewidth]{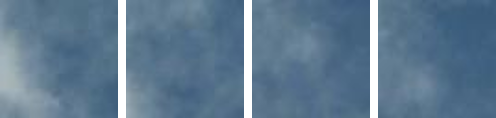}
 }
   \hspace{2mm}
    \subfloat[Synthesis method: NULL; Annotated score: 0]{
     \label{subfig:6450810_AR_s2}
    \includegraphics[height = 0.14\linewidth]{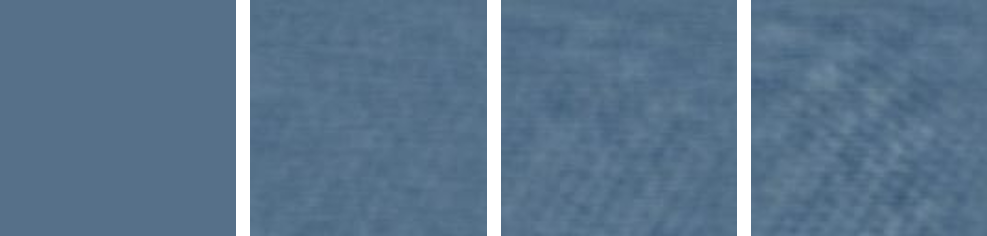}
 }
 \vspace{-3mm}

 }
   \caption{\small Left: the predicted spatial synthesizability scores of SHDT examples; Right: the ‘best’ synthesized results by EDTS methods. In each row, four frames of $64 \times 64$ pixels from the original sequences are shown in the left, and four synthesized frames (expanded to $128 \times 128 $ pixels for spatial synthesis) are shown in the right, along with the annotated spatial synthesizability score. Please check the resulting videos at \url{http://captain.whu.edu.cn/project/DTsynthesizability.html}.}
\label{fig:SHDT_SpatialScore}
\vspace{-3mm}
\end{figure*}

\paragraph{On the spatial synthesizability:}
Fig.~\ref{fig:SHDT_SpatialScore} shows SHDT examples together with their predicted spatial synthesizability. The synthesizability predictor here was trained with all annotated SHDTs except for the test one itself given for prediction. As can be seen, homogeneous, repetitive SHDTs with tiny oscillating dynamics obtain higher scores. The low scores are caused by many factors, such as surface irregularity, uneven illumination, and outliers. In Fig.~\ref{fig:SHDT_SpatialScore}, the ``best'' synthesised dynamic textures by the EDTS methods are also given.
If none of the EDTS methods can synthesize a test example well (``bad'' ones were assigned to ``NULL'' method), synthesis result of a randomly chosen method is shown.
The predicted synthesizability score is more or less consistent with the quality of synthesized DTs. This is crucial because it allows us to select DT regions - also as video parts - that can be synthesized well.

\begin{figure*}[htb!]
\vspace{-3mm}
 \centering
    \subfloat[Synthesizability score: 0.54;]{
     \label{subfig:54pc210}
    \includegraphics[height = 0.095\linewidth]{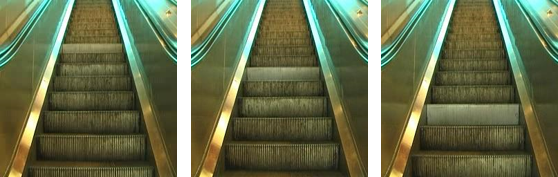}
 }
   \hspace{2mm}
    \subfloat[Synthesis method: LDS; Annotated score: 0.5]{
     \label{subfig:54pc210_256_lds_t2}
    \includegraphics[height = 0.095\linewidth]{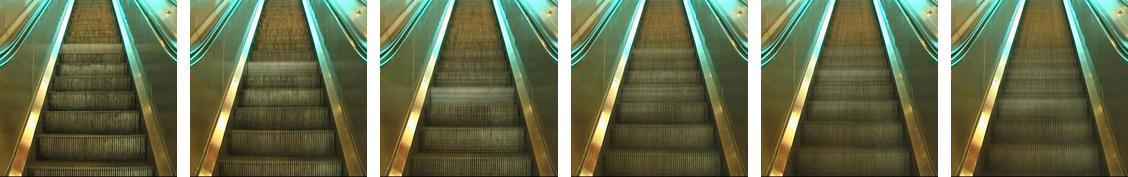}
 }
 \vspace{-3mm}

    \subfloat[Synthesizability score: 0.98;]{
     \label{subfig:64adl10}
    \includegraphics[height = 0.095\linewidth]{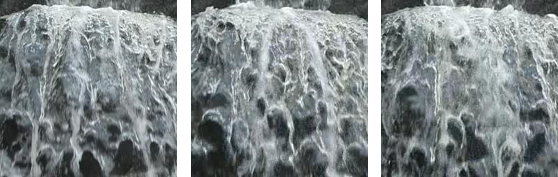}
 }
   \hspace{2mm}
    \subfloat[Synthesis method: Graphcut Textures; Annotated score: 1]{
     \label{subfig:64adl10_256_GC_t2}
    \includegraphics[height = 0.095\linewidth]{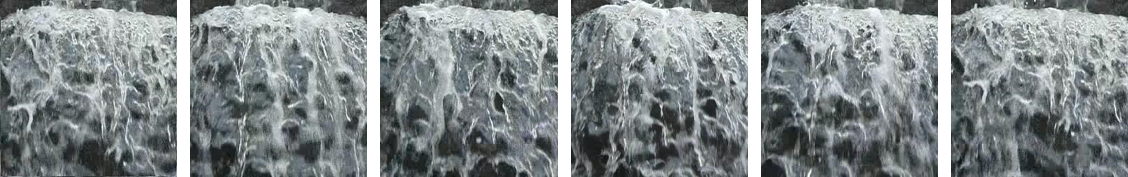}
 }
 \vspace{-3mm}

    \subfloat[Synthesizability score: 0.12;]{
     \label{subfig:647c730}
    \includegraphics[height = 0.095\linewidth]{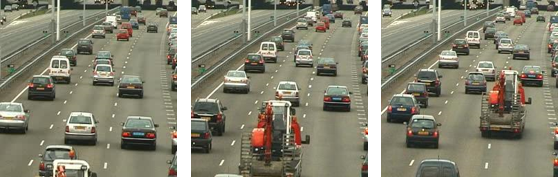}
 }
   \hspace{2mm}
    \subfloat[Synthesis method: NULL; Annotated score: 0]{
     \label{subfig:647c730_256_GC_t2}
    \includegraphics[height = 0.095\linewidth]{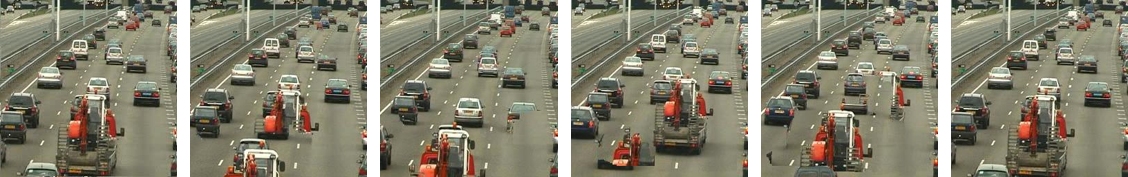}
 }
 \vspace{-3mm}

     \subfloat[Synthesizability score: 0.64;]{
     \label{subfig:55fc310}
    \includegraphics[height = 0.095\linewidth]{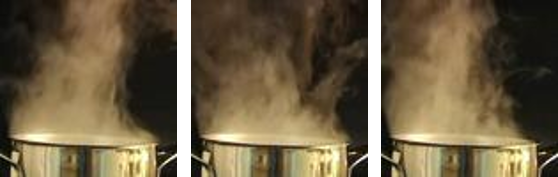}
 }
   \hspace{2mm}
    \subfloat[Synthesis method: STGConvNet; Annotated score: 0.5]{
     \label{subfig:55fc310_100_FCS3_fr50_t1_5}
    \includegraphics[height = 0.095\linewidth]{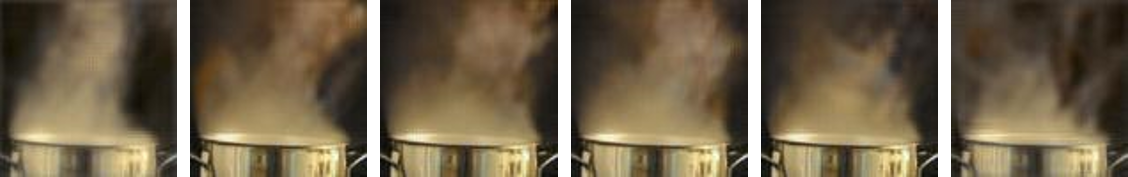}
 }
 \vspace{-3mm}
\caption{\small Left: the predicted temporal synthesizability scores of TDT examples; Right: the ‘best’ synthesized results by EDTS methods. In each row, three frames from the original sequences are shown in the left, and six synthesized frames (extended to twice long as the original for temporal synthesis) are shown in the right, along with the annotated temporal synthesizability score. Please check the resulting videos at \url{http://captain.whu.edu.cn/project/DTsynthesizability.html}.}
\label{fig:TDT_TemporalScore}
\vspace{-3mm}
\end{figure*}

\paragraph{On the temporal synthesizability:}
For temporal synthesizability, Fig.~\ref{fig:TDT_TemporalScore} shows some TDT examples together with their predicted temporal synthesizability. As shown, TDTs with repetitive and slight movement especially for turbulent dynamics of tiny structures obtain higher scores in agreement with the synthesis results. The low quality of temporal synthesis is due to dominant motion of large-structure patterns, time discontinuity and outliers. As can be seen in the 3rd example in Fig.~\ref{fig:TDT_TemporalScore}, artifacts appear in the synthesis of scattered driving cars, where mosaic mismatches are noticeable especially for the large truck.

\subsection{Detect synthesizable regions}
\begin{figure*}[htb!]
 \centering
 \includegraphics[width= 1 \linewidth]{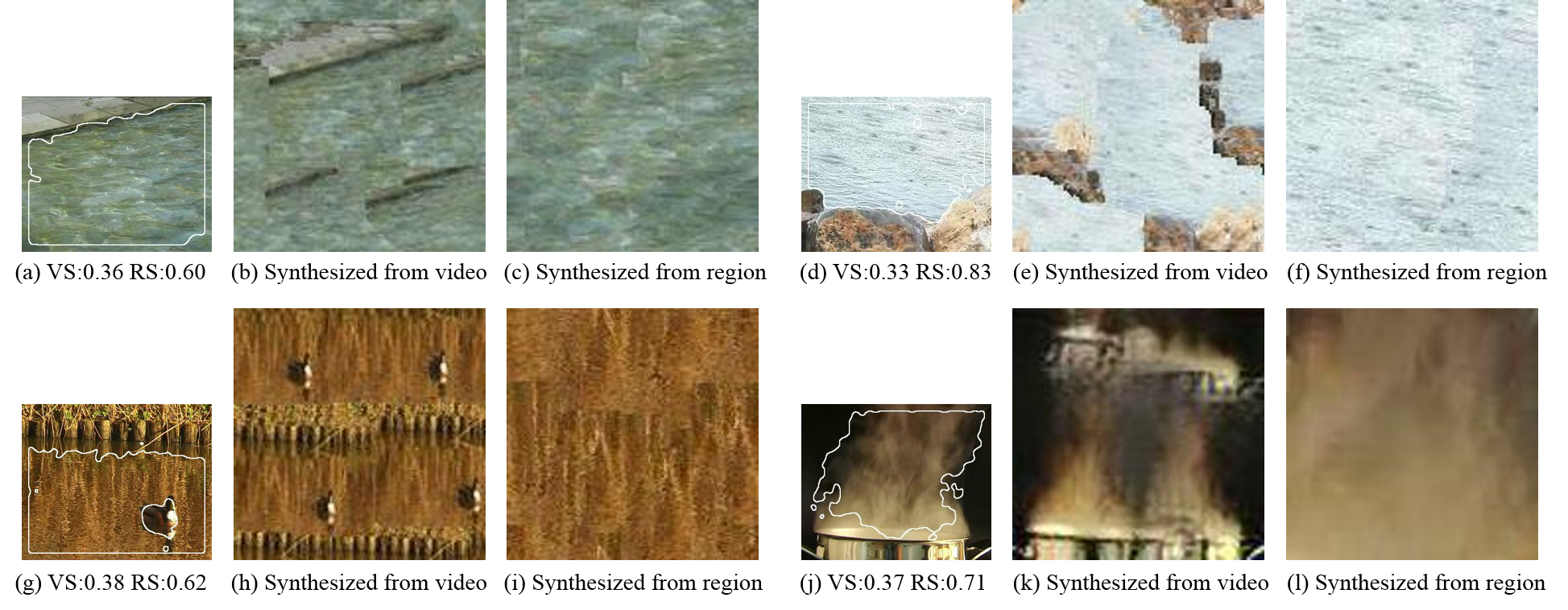}
 \vspace{-5mm}
   \caption{\small Predicted synthesizability and synthesis results for the original videos and detected DT regions: (a) detection in the video; (b) synthesis result of the entire video; (c) synthesis result of the detected region. VS: predicted synthesizability of the entire video; RS: predicted synthesizability of the detected DT region. The same in (d-f), (g-i) and (j-l). Among the synthesis results, (b),(c),(e),(f),(h) and (i) are synthesized by Graphcut Textures, and (k) and (l) by Gatys-DT. Please check the resulting videos at \url{http://captain.whu.edu.cn/project/DTsynthesizability.html}.}
\label{fig:detection_trimming}
\vspace{-3mm}
\end{figure*}

As for detection of the most synthesizable regions, we predict the spatial synthesizability of the original videos and the segmented DT subregions respectively, seen in Fig.~\ref{fig:detection_trimming}. The synthesis results are illustrated, which shows that synthesis is much better for the tailored parts compared to the entire videos. It is thus possible to trim unconstrained videos into synthesizable DT examples. The proposed method also works well for the trimming tasks.

\section{Conclusions}\label{sec:conclude}
This paper investigated the synthesizability of dynamic texture samples via a learning scheme.
To accommodate in more general settings, we proposed a hierarchical architecture to automatically identify dynamic textures from a diverse set of unconstrained videos firstly, followed by partitions of DTs into SHDTs and TDTs according to their spatial modes, and then predict synthesizability scores of DTs to help find good DT examples for synthesis. To this end, we constructed a fairly large dynamic texture dataset and calibrated it in the light of synthesizability. We solved the learning problem by regression models with the proposed SCOP-DT descriptor and other spatiotemporal features. Experimental results show that the proposed hierarchical learning scheme is effective in dynamic texture samples selection, partition and synthesizability prediction. It is helpful to pick out dynamic textures, to find good DT examples for synthesis, to crop synthesizable dynamic texture regions in unconstrained videos, and to suggest an appropriate EDTS method for synthesis. The proposed method of learning synthesizability makes it possible to provide suitable DT examples, thus also facilitating the development of future EDTS methods. For further study, it will be interesting to investigate the relationship between dynamic texture synthesizability and other measures such as  video quality or motion patterns.

\section{Acknowledgement}
We thank Dr. G. Doretto for providing the UCLA datasets.
The authors thank Dr. Gang Liu for his insightful discussion and Prof. Gianfranco Doretto for providing the UCLA datasets.

{\small
\bibliographystyle{ieee}
\bibliography{DT_synthesizability}
}

\end{document}